**Contextualized Medication Information Extraction Using Transformer-based Deep Learning Architectures**

Authors: Aokun Chen, PhD[1,2]
Zehao Yu, MS[1]
Xi Yang, PhD[1]
Yi Guo, PhD[1,2]
Jiang Bian, PhD[1,2]
Yonghui Wu, PhD[1,2]

Affiliation of the authors: [1]Department of Health Outcomes and Biomedical Informatics, College of Medicine, University of Florida, Gainesville, Florida, USA
[2]Cancer Informatics Shared Resource, University of Florida Health Cancer Center, Gainesville, Florida, USA

Corresponding author: Yonghui Wu, PhD
Clinical and Translational Research Building
2004 Mowry Road, PO Box 100177
Gainesville, FL, USA, 32610
Phone: 352-294-8436
Email: yonghui.wu@ufl.edu

Word count: 3,266

# ABSTRACT

**Objective**

To develop a natural language processing (NLP) system to extract medications and contextual information that help understand drug changes. This project is part of the 2022 n2c2 challenge.

**Materials and methods**

We developed NLP systems for medication mention extraction, event classification (indicating medication changes discussed or not), and context classification to classify medication changes context into 5 orthogonal dimensions related to drug changes. We explored 6 state-of-the-art pretrained transformer models for the three subtasks, including GatorTron, a large language model pretrained using >90 billion words of text (including >80 billion words from >290 million clinical notes identified at the University of Florida Health). We evaluated our NLP systems using annotated data and evaluation scripts provided by the 2022 n2c2 organizers.

**Results**

Our GatorTron models achieved the best F1-scores of 0.9828 for medication extraction (ranked 3rd), 0.9379 for event classification (ranked 2nd), and the best micro-average accuracy of 0.9126 for context classification. GatorTron outperformed existing transformer models pretrained using smaller general English text and clinical text corpora, indicating the advantage of large language models.

**Conclusion**

This study demonstrated the advantage of using large transformer models for contextual medication information extraction from clinical narratives.

**INTRODUCTION**

The exposure and changes of medications are critical information to study health outcomes and pharmaceutical outcomes. Much details of medication changes (e.g., start, stop) and the corresponding contextualized information (e.g., negation and temporality) are usually captured as free text in clinical narratives, which cannot be directly used in clinical studies that require structured data fields. [1] Therefore, studies that solely rely on structure EHR for medication records is jeopardized by the system error from computerized provider order entry (COPE) systems. [2] Previous studies have applied natural language processing (NLP) to extract medication names, attributes (e.g., dose, frequency), and potential adverse drug events (ADEs), yet it is still challenging to extract comprehensive contextualized medication information for medication changes from clinical narratives.

NLP is evolving fast in the clinical domain as the breakthrough of deep learning-based machine learning algorithms and the power of large language models (LLMs) pretrained using large-scale electronic health records (EHRs). The 2022 National NLP Clinical Challenge (n2c2) organized an open challenge with shared tasks focusing on the extraction of medications and contextualized information including negation, temporality (e.g., past, present), certainty (e.g., hypothetical, conditional), and actor (e.g., patient, physician). The challenge consists of three subtasks: (1) medication extraction – extract medication mentions; (2) event classification – classify medication mentions into disposition (medication change discussed), no-disposition (no change discussed), or undetermined (need more information); and (3) context classification – classify the medication mentions of the 'disposition' group into 5 dimensions including Action (e.g. start, stop), Negation (e.g. negated), Temporality (e.g. past, present), Certainty (e.g. hypothetical, conditional), and

Actor (e.g. patient, physician). We explored 6 pretrained transformer models from the general English domain and clinical domain for the 3 subtasks and our NLP systems were ranked 3rd in subtask 1 and ranked 2nd in subtask 2.

## BACKGROUND

Researchers have explored NLP methods for medication information extraction from clinical narratives using benchmark datasets contributed by a series of open challenges organized in the clinical NLP community. In 2009, the third i2b2 Workshop organized an open challenge focusing on the identification of medications, their dosages, routes of administration, frequencies, durations, and reasons for administration from discharge summaries. [3] The 2010 [4] and 2012 [5] i2b2 challenges also have shared tasks focusing on the extraction of medications as part of treatment concepts and further explored a few contextual information such as Temporality and Negation. Later, the 2018 NLP challenge [6] for detecting Medication and Adverse Drug Events from electronic health records (MADE 1.0) and 2018 n2c2 challenge [7] focused on the detection of medication and medication-induced ADEs, both of which have subtasks for medication information extraction. These open challenges provided valuable benchmark datasets publicly available through data use agreements, which greatly advanced the development of NLP systems for medication information extraction. We have participated in these challenges and developed machine learning-based NLP systems among the top teams [6,8,9].

Various NLP approaches for medication information extraction, including dictionary-based, rule-based, and machine learning-based, have been developed. Early medication extraction systems are often rule-based systems that utilized dictionaries of medications. For example, seven out of the top ten teams in the 2009 i2b2 challenge submitted rule-based solutions. The top 10 teams achieved exact F1 scores ranging from 0.764 to 0.857. [4] Xu *et al.* developed MedEx [1,10], one of the most widely used medication information extraction systems for clinical narratives using semantic parsing based on a context-free grammar and a comprehensive lexicon generated from many resources such as the RXNorm. Later, more and more systems applied machine learning solutions such as conditional random fields (CRFs) [11] and Hidden Markov Model (HMM) [12]. With the breakthrough of deep learning, a family of machine learning models based on deep neural network (DNN) architectures, recent studies have explored various DNNs for medical information extraction. [13] For example, we have applied convolutional neural networks (CNNs) to extract medications from clinical notes in Chinese [14]; Wei *et al.* [15] and Jagannatha *et al.* [16] explored a recurrent neural network (RNN) implemented using bi-directional Long-Short-Term-Memory (Bi-LSTM) for medication and ADE detection from clinical notes. In the 2018 n2c2 challenge on medications and ADEs, nine out of ten teams adopted the Bi-LSTM-CRF model. The best system achieved a lenient micro-averaged F1 score of 94.18 while the median F1 score for all systems was 0.9052 [7].

Recently, deep learning models based on the transformer architecture have become state-of-the-art solutions for many NLP tasks. [17] The original transformer was built in an encoder-decoder structure using the self-attention mechanism. Compared with CNNs, the transformer reduced the complexity of the model and enables the learning of relations between distant features. Transformer-based NLP models have achieved state-of-the-art performance for almost all NLP tasks such as named entity recognition, relation extraction, natural language inference, and question answering. Bidirectional Encoder Representations from Transformers (BERT) [18] and RoBERTa [19] are two popular transformer architectures. Transformer-based deep learning models split the learning into pretraining - where LLMs are pretrained using large-scale unlabeled corpora, and fine-tuning - where the pretrained LLMs are fine-tuned using a small set of data with human annotations. Thus, one pretrained LLM can be applied to solve many downstream NLP tasks. Previously we have developed transformer models for clinical concept extraction including medications. [20]

In this study, we developed transformer-based NLP models to extract medication mentions and contextualized information using the benchmark dataset provided by the 2022 n2c2 challenge. We explored 4 pretrained transformer models from the biomedical domain and two new transformer models pretrained using UF Health clinical text, including GatorTron [21] – an LLM pretrained using >90 billion words of text, and GatorTronS [22] – an LLM pretrained using >20 billion words of synthetic clinical text generated using a GPT-3 [23] based generative clinical language model - GatorTronGPT [24]. Our system was ranked 3$^{rd}$ in subtask 1 and 2$^{nd}$ in subtask 2 according to the official evaluation results.

**METHODS**

**Dataset**

The 2022 n2c2 challenge organizers developed a corpus of 500 de-identified clinical notes from the Partners Healthcare's CMED (Contextualized Medication Event Dataset). [25] Annotators manually annotated medications, events, and 6 dimensions of contextualized medication information. The corpus was divided into a training set of 350 notes, a validation set of 50 notes and a test set of 100 notes. Table 1 provides detailed statistics for the training and test sets. (We combined the validation set with training for simplicity)

Table 1. Summary of events

| Task | Categories | Count in train (%) | Count in test (%) | Task | Categories | Count in train (%) | Count in test (%) |
|---|---|---|---|---|---|---|---|
| Event | No Disposition | 5260 (72.8%) | 1326 (74.4%) | Temporality | Past | 744 (52.7%) | 173 (51.6%) |
| | Disposition | 1412 (19.5%) | 335 (18.9%) | | Present | 494 (35.0%) | 132 (39.4%) |
| | Undetermined | 557 (7.7%) | 122 (6.8%) | | Future | 145 (10.3%) | 29 (8.7%) |
| Action | Start | 568 (40.2%) | 131 (39.1%) | | Unknown | 29 (2.1%) | 1 (0.3%) |
| | Stop | 340 (24.1%) | 67 (20.0%) | Certainty | Certain | 1176 (83.3%) | 281 (83.9%) |
| | Increase | 129 (9.1%) | 22 (6.6%) | | Hypothetical | 134 (9.5%) | 33 (9.9%) |
| | Decrease | 54 (3.8%) | 13 (3.9%) | | Conditional | 100 (7.1%) | 15 (4.5%) |
| | Unique Dose | 285 (20.2%) | 88 (26.3%) | | Unknown | 2 (0.1%) | 6 (1.8%) |
| | Other Change | 1 (0.1%) | 0 (0.0%) | Actor | Physician | 1278 (90.5%) | 311 (92.8%) |
| | Unknown | 35 (2.5%) | 14 (4.2%) | | Patient | 106 (7.5%) | 17 (5.1%) |
| Negation | Negated | 32 (2.3%) | 6 (1.8%) | | Unknown | 28 (2.0%) | 7 (2.1%) |
| | Not Negated | 1380 (97.7%) | 329 (98.2%) | Medication extraction | Medication | 7229 (100%) | 1783 (100%) |

**Preprocessing**

We reused the preprocessing pipelines developed in our previous study [20] to perform tokenization, sentence boundary detection, and 'BIO' format transformation. As different transformer models applied different word segmentation algorithms, our preprocessing module dynamically applied word segmentation algorithms according to the transformer model and aligned the word-level 'BIO' tags to the subtoken-level 'BIO' tags. The detailed preprocessing algorithm can be accessed from our GitHub repository: https://github.com/uf-hobi-informatics-lab/NLPreprocessing.

**Medication extraction**

We approached medication extraction as a named entity recognition (NER) task and applied transformer-based deep learning methods. We adopted the standard 'BIO' format to represent the medication concept. Then, transformer-based deep learning models were used to classify words into three categories of labels (B, I, or O). Using pretrained transformer models, we generated distributed representations and used a classification layer (a linear layer with softmax activation) to calculate a probability score for each 'BIO' category. The cross-entropy loss was used for optimization.

**Event and context classification**

We approached event and contextualized medication information as text classification tasks and developed transformer-based classifiers. Specifically, we identified sentences containing medications and applied pretrained transformer models to generate sentence-level representation (e.g., the [CLS] token in BERT) and concept-level representation (e.g., the [S] and [E] tokens in BERT). A maximum length of 256 tokens was used for the input sentences. Sentences containing more than 256 tokens were truncated. We concatenated the sentence-level representation and concept-level representation into a classification layer to calculate a probability for each of the two categories: disposition and non-disposition. The cross-entropy loss was used for fine-tuning. For context classification, we applied the same strategy to further classify medications of the

'disposition' group into 5 orthogonal dimensions: Action (start, stop), Negation (negated, not negated), Temporality (past, present), Certainty (hypothetical, conditional), and Actor (patient, physician). We trained individual classifiers for each dimension and aggregated the results using a post-processing.

**Transformer-based machine learning models**

We explored 2 pretrained transformer models from the general English domain, including Roberta and ALBERT; and 4 pretrained transformers from the clinical domain, including Roberta_MIMIC, ALBERT_MIMIC, GatorTron, and GatorTronS. We did not include the original BERT model as both GatorTron and GatorTronS were implemented using the same BERT-based architecture and pretrained using a much larger corpus. [21] Our previous study shows that GatorTron outperformed other BERT-based transformers including BioBERT and ClinicalBERT for medication extraction. [20,21,26,27]

*ALBERT and ALBERT_MIMIC*

Lan *et al.* developed A Lite BERT (ALBERT) for self-supervised learning of language representations [28]. Compared with the original BERT model, ALBERT adopted factorized embedding parameterization and cross-layered parameter sharing with the self-supervised loss for sentence-order prediction. We adopted the ALBERT model implemented in the Huggingface with 128M parameters. [29]

*RoBERTa and RoBERTa_MIMIC*

RoBERTa is an optimized BERT model developed by Liu *et al.* [19]. RoBERTa introduced new strategies including dynamic masking, full sentence sampling, large mini-batches, large byte level encoding, and removed next sentence prediction loss. RoBERTa MIMIC utilized the same optimization of RoBERTa but trained over the MIMIC data set. We explored the RoBERTa model implemented in the Huggingface with 355M parameters. [30]

*GatorTron and GatorTronS*

GatorTron is a BERT-style LLM pretrained using >90 billion words of text. [21] GatorTronS is also a BERT-style LLM pretrained using >20 billion words of synthetic clinical text generated using a GPT-3 model, GatorTronGPT. [22] We used the version with 345 million parameters for both GatorTron and GatorTronS.

**Training strategies**

For medication extraction, we followed the standard NER training procedure to fine-tune transformer models to recognize mediations using the training and validation sets provided in this challenge. Specifically, we train models using the training set of 350 notes and monitoring the performance using the validation set of 50 notes. For each transformer model, the best model based on the validation performance was submitted. For event and context classification, we followed the similar procedure to fine-tune transformer models for classification using the training and validation sets. The best classification model based on the validation results were submitted.

**The end-to-end system**

We integrated the medication extraction, event classification, and the context classification into a unified pipeline for the end-to-end task. The best models based on the validation performance were selected for submission.

**Experiment and evaluation**

We reused the pretrained models from the public GitHub repository for two transformer models from the general domain, including RoBERTa, and ALBERT. For the two clinical transformer models, we adopted the RoBERTa_MIMIC and ALBERT_MIMIC model developed by fine-tuning the general models using clinical text from the MIMIC III database with > 745 million words in our previous study [20]. GatorTron and GatorTronS models were developed by training from scratch using >90 billion words of text (including >82 billion words of de-identified clinical text from UF Health) in our previous studies. [21] Following the evaluation metrics used in this challenge, the micro- and macro- averaged precision, recall, and F1-score were used to evaluate medication extraction; the micro-averaged precision, recall and F1-score were used to evaluate the event classification and the contextual classification. All evaluation scores were calculated using the official evaluation script and the test data set provided in this challenge.

**RESULTS**

**Table 2** shows performance of medication extraction for all 6 transformer models. GatorTron achieved the best micro-average F1-score of 0.9828 (ranked 3[rd] in this challenge), followed by RoBERTa_MIMIC (0.9801) and GatorTronS (0.9791). GatorTronS achieved the best Macro-average F1-score of 0.9672, followed by GatorTron (0.9659) and RoBERTa_MIMIC (0.9643).

All clinical transformers outperformed the transformer models trained using general English text with large margins.

Table 2. Performance of medication extraction on the test set.

| Model | Micro | | | Macro | | |
|---|---|---|---|---|---|---|
| | Precision | Recall | F1 | Precision | Recall | F1 |
| GatorTron | 0.9772 | **0.9887** | **0.9828** | 0.9602 | **0.9717** | 0.9659 |
| GatorTronS | **0.9840** | 0.9743 | 0.9791 | **0.9720** | 0.9624 | **0.9672** |
| RoBERTa | 0.8746 | 0.8772 | 0.8759 | 0.8588 | 0.8614 | 0.8601 |
| RoBERTa MIMIC | 0.9832 | 0.9801 | 0.9816 | 0.9703 | 0.9584 | 0.9643 |
| ALBERT | 0.8472 | 0.8673 | 0.8571 | 0.8221 | 0.8416 | 0.8317 |
| ALBERT MIMIC | 0.9673 | 0.9653 | 0.9663 | 0.9355 | 0.9337 | 0.9346 |

Best Micro and Macro precision, recall, and F1-scores are highlighted in bold.

**Table 3** compares the 6 transformer models for event classification using the test set. GatorTron achieved the best overall micro-average score of 0.9379 (ranked 2[nd] in this challenge) and the best F1 scores (0.8726, 0.9652, and 0.6967) for all three categories. Two new transformer models pretrained using a larger corpus (i.e., GatorTron and GatorTronS) outperformed 4 existing transformer models pretrained using a smaller corpus (RoBERTa, RoBERTa_MIMIC, ALBERT, ALBERT_MIMIC).

Table 3. Performance of event classification for the test set.

| Model | Overall (Micro) | Disposition (Strict) | | | No Disposition (Strict) | | | Undetermined (Strict) | | |
|---|---|---|---|---|---|---|---|---|---|---|
| | | Pre | Rec | F1 | Pre | Rec | F1 | Pre | Rec | F1 |
| GatorTron | **0.9379** | **0.8782** | 0.8671 | **0.8726** | 0.9648 | 0.9655 | 0.9652 | 0.6911 | **0.7025** | **0.6967** |
| GatorTronS | 0.9362 | 0.8490 | 0.8232 | 0.8359 | 0.8893 | 0.9310 | 0.9097 | 0.7258 | 0.5172 | 0.6040 |
| RoBERTa | 0.8588 | 0.8111 | 0.7374 | 0.7725 | 0.8346 | 0.9255 | 0.8777 | 0.6600 | 0.3793 | 0.4818 |
| RoBERTa MIMIC | 0.9251 | 0.8323 | **0.8797** | 0.8554 | 0.9646 | 0.9609 | 0.9628 | **0.7383** | 0.6529 | 0.6930 |
| ALBERT | 0.8472 | 0.8111 | 0.7374 | 0.7725 | 0.8346 | 0.9255 | 0.8777 | 0.6600 | 0.3793 | 0.4818 |
| ALBERT MIMIC | 0.9179 | 0.8012 | 0.8797 | 0.8386 | 0.9666 | 0.9533 | 0.9599 | 0.7103 | 0.6281 | 0.6667 |

Best precision, recall, and F1-scores are highlighted in bold.

Table 4. Performance of context classification on the test set.

| Model | Accuracy | | | | | |
|---|---|---|---|---|---|---|
| | Overall | Action | Negation | Temporal | Certainty | Actor |
| GatorTron | **0.9126** | **0.8862** | **0.9790** | 0.8503 | **0.9102** | 0.9371 |
| GatorTronS | 0.9080 | 0.8503 | **0.9790** | 0.8683 | 0.9051 | 0.9371 |
| RoBERTa | 0.8417 | 0.7994 | 0.9740 | 0.7198 | 0.7994 | 0.9158 |
| RoBERTa MIMIC | 0.9121 | 0.8729 | 0.9740 | **0.8774** | 0.9096 | **0.9303** |
| ALBERT | 0.8196 | 0.6064 | 0.9740 | 0.8025 | 0.7994 | 0.9158 |
| ALBERT MIMIC | 0.9072 | 0.8453 | 0.9740 | 0.8882 | 0.8943 | 0.9342 |

Best accuracies are highlighted in bold.

Table 4 compares the 6 transformers for context classification using the test set. GatorTron achieved the best overall accuracy of 0.9126, outperforming other transformer models. The end-to-end system applied the GatorTron model for all three subtasks consecutively and achieved an overall accuracy of 0.6178.

**DISCUSSION AND CONCLUSION**

Understanding the exposure and changes of medications are critical in assessing various health outcomes and pharmaceutical outcomes using EHRs. The 2022 n2c2 challenge was organized to examine state-of-the-art NLP systems to extract medication mentions and determine the contextual categories indicating drug changes. We participated in all three subtasks and developed transformer-based solutions using 6 pretrained transformer models. Our systems achieved the third-best performance (micro-average F1 score of 0.9828) for subtask 1 and achieved the second-best performance for subtask 2 (overall micro accuracy of 0.9379). The experimental results show that our GatorTron models trained using >90 billion words of text outperformed existing pretrained language models for medication extraction and event/context classification. Our NLP systems can be applied to help characterize mediation changes to better study health outcomes and pharmaceutical outcomes.

For medication extraction, our GatorTron model achieved the best F1-score of 0.9828, indicating the efficiency of transformer-based LLMs. Among the 6 transformer models, those pretrained using clinical text outperformed others pretrained using general English text (i.e., RoBERTa and ALBERT), which is consistent with previous studies reporting that domain-specific clinical transformers outperformed general transformer models on clinical concept extraction. [20] Among the clinical transformers, the GatorTron model trained using >90 billion words of text achieved the best scores outperforming other clinical transformers pretrained using a much smaller clinical corpus. GatorTron models improved the performance of medication extraction mainly on recall, indicating that LLMs captured new documenting patterns from a much larger corpus. This observation is consistent with our previous study that scaling up data size improved various clinical NLP tasks. [21] RoBERTa_MIMIC showed comparable results with less pretraining data, indicating that the benefit of LLMs to concept extraction and event classification is moderate, which is consistent with the observation reported in our previous study.[21]

For event classification, our GatorTron models also achieved the best micro-average accuracy (0.9379) among all categories. GatorTronS achieved performance comparable to GatorTron (0.35% difference). Compared with other model, GatorTron achieved better performance (e.g., 1.28% higher than RoBERTa). In this challenge, GatorTron ranked the third for event classification, which is 0.82% lower than the best performed model. Compared with medication extraction (a phrase-level NLP task), the performance improvements derived from GatorTron and GatorTronS are remarkably larger for event classification (a sentence-level task). Our finding suggests that larger transformer models benefit more for complex NLP tasks require long pieces of text. When looking into the scores of each category, we observed that all transformer models had remarkably lower scores for the 'undetermined' category. We examined the distribution of the 'undetermined' category and found that this category has a lower proportion in the test set than in the training set (train vs. test = 7.7% vs. 6.8%). Similar results were observed in context classification, where better performances were achieved for Actor and Negation, where there were similar distributions in the test set compared with the training set. This suggests that unbalanced data distribution is still a challenge. Our systems did not handle well instances with contradicting labels for one medication when the labels in the five dimensions of context are not mutually exclusive. For example, the start and stop events of the same medication may be discussed in different sections of a single clinical note. Future studies should explore solutions for these samples with contradicting labels.

Transformer-based NLP models achieved good performance for single subtask, yet the end-to-end system achieved a much lower performance (overall accuracy of 0.6178), indicating that extracting comprehensive contextual medication information with multiple dimensions is still a challenging task. Further studies should explore algorithms that could alleviate the unbalanced distribution of samples and improve the performance of extracting multi-dimensional contextual medication information to understand drug changes. The experimental results show that GatorTronS, an LLM trained using synthetic clinical text, achieved performances comparable to GatorTron, an LLM trained using real-world clinical text, for medication extraction (0.9791 vs 0.9828), event classification (0.9362 vs 0.9379), and context classification (0.9080 vs 0.9126). Our findings support the potential utility of synthetic text generation from generative clinical LLMs such as GatorTronGPT to fill the gap in accessing large-scale clinical text and sharing clinical NLP models. Future studies should examine synthetic text generation of generative LLMs for other NLP applications.

## DATA AVAILIABILITY

The datasets are available from the n2c2 challenge website with a data use agreement.

The computer codes for subtask 1 medication extraction can be accessed from: https://github.com/uf-hobi-informatics-lab/ClinicalTransformerNER.

The computer codes for subtasks 2 and 3 on context classification can be accessed from: https://github.com/uf-hobi-informatics-lab/ClinicalTransformerClassification.

## ACKNOWLEDGMENTS

We would like to thank the n2c2 organizers to provide the annotated corpus and the guidance for this challenge. We gratefully acknowledge the support of NVIDIA Corporation with the donation of the GPUs used for this research.

## FUNDING STATEMENT

This study was partially supported by grants from the Patient-Centered Outcomes Research Institute® (PCORI®) (ME-2018C3-14754), the National Institute on Aging (1R56AG069880, R21AG068717), the National Cancer Institute (1R01CA246418, 3R01CA246418-02S1,

## COMPETING INTERESTS STATEMENT

Aokun Chen, Zehao Yu, Xi Yang, Yi Guo, Jiang Bian, and Yonghui Wu have no conflicts of interest that are directly relevant to the content of this study.

## CONTRIBUTORSHIP STATEMENT

AC, XY, and YW were responsible for the overall design, development, and evaluation of this study. AC, XY, and ZY developed the NLP systems. AC and YW did the bulk of the writing, YG and JB also contributed to writing and editing of this manuscript. All authors reviewed the manuscript critically for scientific content, and all authors gave final approval of the manuscript for publication.

## SUPPLEMENTARY MATERIAL

None.

## REFERENCES

1   Xu H, Stenner SP, Doan S, *et al.* MedEx: a medication information extraction system for clinical narratives. *Journal of the American Medical Informatics Association* 2010;**17**:19–24. doi:10.1197/jamia.M3378

2   Kinlay M, Zheng WY, Burke R, *et al.* Medication errors related to computerized provider order entry systems in hospitals and how they change over time: A narrative review. *Res Social Adm Pharm* 2021;**17**:1546–52. doi:10.1016/j.sapharm.2020.12.004

3   Uzuner Ö, Solti I, Cadag E. Extracting medication information from clinical text. *Journal of the American Medical Informatics Association* 2010;**17**:514–8. doi:10.1136/jamia.2010.003947

4   Uzuner Ö, South BR, Shen S, *et al.* 2010 i2b2/VA challenge on concepts, assertions, and relations in clinical text. *Journal of the American Medical Informatics Association* 2011;**18**:552–6. doi:10.1136/amiajnl-2011-000203

5   Sun W, Rumshisky A, Uzuner O. Evaluating temporal relations in clinical text: 2012 i2b2 Challenge. *Journal of the American Medical Informatics Association* 2013;**20**:806–13. doi:10.1136/amiajnl-2013-001628

6   Yang X, Bian J, Gong Y, *et al.* MADEx: A System for Detecting Medications, Adverse Drug Events, and Their Relations from Clinical Notes. *Drug Saf* 2019;**42**:123–33. doi:10.1007/s40264-018-0761-0

7   Henry S, Buchan K, Filannino M, *et al.* 2018 n2c2 shared task on adverse drug events and medication extraction in electronic health records. *Journal of the American Medical Informatics Association* 2020;**27**:3–12. doi:10.1093/jamia/ocz166

8   Tang B, Wu Y, Jiang M, *et al.* A hybrid system for temporal information extraction from clinical text. *Journal of the American Medical Informatics Association* 2013;**20**:828–35. doi:10.1136/amiajnl-2013-001635

9   Yang X, Bian J, Fang R, *et al.* Identifying relations of medications with adverse drug events using recurrent convolutional neural networks and gradient boosting. *J Am Med Inform Assoc* 2019;**27**:65–72. doi:10.1093/jamia/ocz144

10  Jiang M, Wu Y, Shah A, *et al.* Extracting and standardizing medication information in clinical text – the MedEx-UIMA system. *AMIA Jt Summits Transl Sci Proc* 2014;**2014**:37–42.

11  Lafferty JD, McCallum A, Pereira FCN. Conditional Random Fields: Probabilistic Models for Segmenting and Labeling Sequence Data. In: *Proceedings of the Eighteenth International Conference on Machine Learning*. San Francisco, CA, USA: : Morgan Kaufmann Publishers Inc. 2001. 282–9.

12  de Bruijn B, Cherry C, Kiritchenko S, *et al.* Machine-learned solutions for three stages of clinical information extraction: the state of the art at i2b2 2010. *J Am Med Inform Assoc* 2011;**18**:557–62. doi:10.1136/amiajnl-2011-000150

13  Hahn U, Oleynik M. Medical Information Extraction in the Age of Deep Learning. *Yearb Med Inform* 2020;**29**:208–20. doi:10.1055/s-0040-1702001

14  Wu Y, Jiang M, Lei J, *et al.* Named Entity Recognition in Chinese Clinical Text Using Deep Neural Network. *Stud Health Technol Inform* 2015;**216**:624–8.

15  Wei Q, Ji Z, Li Z, *et al.* A study of deep learning approaches for medication and adverse drug event extraction from clinical text. *J Am Med Inform Assoc* 2019;**27**:13–21. doi:10.1093/jamia/ocz063

16  Jagannatha AN, Yu H. Bidirectional RNN for Medical Event Detection in Electronic Health Records. *Proc Conf* 2016;**2016**:473–82.

17  Vaswani A, Shazeer N, Parmar N, *et al.* Attention Is All You Need. 2017. doi:10.48550/arXiv.1706.03762

18  Devlin J, Chang M-W, Lee K, *et al.* BERT: Pre-training of Deep Bidirectional Transformers for Language Understanding. 2019. doi:10.48550/arXiv.1810.04805

19  Liu Y, Ott M, Goyal N, *et al.* RoBERTa: A Robustly Optimized BERT Pretraining Approach. 2019. doi:10.48550/arXiv.1907.11692

20  Yang X, Bian J, Hogan WR, *et al.* Clinical concept extraction using transformers. *Journal of the American Medical Informatics Association* 2020;**27**:1935–42. doi:10.1093/jamia/ocaa189

21  Yang X, Chen A, PourNejatian N, *et al.* A large language model for electronic health records. *npj Digit Med* 2022;**5**:1–9. doi:10.1038/s41746-022-00742-2

22  GatorTron-S | NVIDIA NGC. NVIDIA NGC Catalog. https://catalog.ngc.nvidia.com/orgs/nvidia/teams/clara/models/gatortron_s (accessed 6 Jan 2023).

23  GPT-3: Its Nature, Scope, Limits, and Consequences | Minds and Machines. https://dl.acm.org/doi/10.1007/s11023-020-09548-1 (accessed 6 Jan 2023).

24  SynGatorTron: A Large Clinical Natural Language Generation Model for Synthetic Data Generation and Zero-shot Tasks | NVIDIA On-Demand. NVIDIA. https://www.nvidia.com/en-us/on-demand/session/gtcspring22-s41638/ (accessed 10 Mar 2023).

25  Mahajan D, Liang JJ, Tsou C-H. Toward Understanding Clinical Context of Medication Change Events in Clinical Narratives. *AMIA Annu Symp Proc* 2022;**2021**:833–42.

26  Lee J, Yoon W, Kim S, *et al.* BioBERT: a pre-trained biomedical language representation model for biomedical text mining. *Bioinformatics* 2020;**36**:1234–40. doi:10.1093/bioinformatics/btz682

27  Alsentzer E, Murphy J, Boag W, *et al.* Publicly Available Clinical BERT Embeddings. In: *Proceedings of the 2nd Clinical Natural Language Processing Workshop*. Minneapolis, Minnesota, USA: : Association for Computational Linguistics 2019. 72–8. doi:10.18653/v1/W19-1909

28  Lan Z, Chen M, Goodman S, *et al.* ALBERT: A Lite BERT for Self-supervised Learning of Language Representations. 2020. doi:10.48550/arXiv.1909.11942

29  ALBERT. https://huggingface.co/docs/transformers/model_doc/albert (accessed 6 Jan 2023).

30  RoBERTa. https://huggingface.co/docs/transformers/model_doc/roberta (accessed 6 Jan 2023).